\documentclass{article}

\usepackage{bbm,bm}
\usepackage{epsfig,ifthen}
\usepackage{latexsym}
\usepackage{amsfonts}
\usepackage{eepic}
\usepackage{amsopn}
\usepackage{amsmath,amssymb,amsthm,rotating,graphicx,rotating,float}
\usepackage{accents}
\usepackage[backref,pageanchor=true,plainpages=false,
pdfpagelabels,bookmarks,bookmarksnumbered,breaklinks,
pdfborder={0 0 0},  
]{hyperref}
\usepackage[mathscr]{eucal}
\usepackage{url}
\usepackage{graphicx}
\usepackage{csquotes}
\usepackage{verbatim}
\usepackage{ulem}
\usepackage{color} 
\usepackage{natbib}

\usepackage{algorithm}
\usepackage{algorithmic}

\usepackage{hyperref}

\usepackage[accepted]{whi2017}

\icmltitlerunning{A Formal Framework to Characterize Interpretability of Procedures}
\newcommand{\eat}[1]{}
\newtheorem{definition}{Definition}[section]

\newlength{\dhatheight}

\newcommand{\ignore}[1]{}
\newcommand{\todo}[1]{}
\newcommand{\oldstuff}[1]{}

\newsavebox{\savepar}
\newenvironment{bigboxit}{\begin{center}\begin{lrbox}{\savepar}
\begin{minipage}[h]{3.15in}
\normalfont
\begin{flushleft}}
{\end{flushleft}\end{minipage}\end{lrbox}\fbox{\usebox{\savepar}}
\end{center}}

\makeatletter
\newcommand{\vast}{\bBigg@{3}}
\newcommand{\Vast}{\bBigg@{4}}
\makeatother

\begin{document} 

\twocolumn[
\icmltitle{A Formal Framework to Characterize Interpretability of Procedures}



\icmlsetsymbol{equal}{*}

\begin{icmlauthorlist}
\icmlauthor{Amit Dhurandhar}{to}
\icmlauthor{Vijay Iyengar}{to}
\icmlauthor{Ronny Luss}{to}
\icmlauthor{Karthikeyan Shanmugam}{to}
\end{icmlauthorlist}

\icmlaffiliation{to}{IBM Research, Yorktown Heights, NY}

\icmlcorrespondingauthor{Amit Dhurandhar}{adhuran@us.ibm.com}

\icmlkeywords{interpretability, formalizing}

\vskip 0.3in
]



\printAffiliationsAndNotice{}  

\begin{abstract} 
We provide a novel notion of what it means to be interpretable, looking past the usual association with human understanding. Our key insight is that interpretability is not an absolute concept and so we define it relative to a target model, which may or may not be a human. We define a framework that allows for comparing interpretable procedures by linking it to important practical aspects such as accuracy and robustness. We characterize many of the current state-of-the-art interpretable methods in our framework portraying its general applicability.
\end{abstract} 

\section{Introduction}
\label{intro}
What does it mean for a model to be interpretable? From our human perspective, interpretability typically means that the model can be explained, a quality which is imperative in almost all real applications where a human is responsible for consequences of the model. However good a model might have performed on historical data, in critical applications, interpretability is necessary to justify, improve, and sometimes simplify decision making.

A great example of this is a malware detection neural network \cite{malwarecom} which was trained to distinguish regular code from malware. The neural network had excellent performance, presumably due to the deep architecture capturing some complex phenomenon opaque to humans, but it was later found that the primary distinguishing characteristic was the grammatical coherance of comments in the code, which were either missing written poorly in the malware as opposed to regular code. In hindsight, this seems obvious as you wouldn't expect someone writing malware to expend effort in making it readable. This example shows how the interpretation of a seemingly complex model can aid in creating a simple rule.

The above example defines interpretability as humans typically do: we require the model to be understandable. This thinking would lead us to believe that, in general, complex models such as random forests or even deep neural networks are not interpretable. However, just because we cannot always understand what the complex model is doing does not necessarily mean that the model is not interpretable in some other useful sense. It is in this spirit that we define the novel notion of $\delta$-interpretability that is more general than being just relative to a human. 

We offer an example from the healthcare domain \cite{chang2010}, where interpretability is a critical modeling aspect, as a running example in our paper. The task is predicting future costs based on demographics and past insurance claims (including doctor visit costs, justifications, and diagnoses) for members of the population. 
The data used in \cite{chang2010} represents diagnoses using ICD-9-CM (International Classification
of Diseases) coding which had on the order of 15,000 distinct codes at the time of the study. The high dimensional nature of diagnoses led to the development of various abstractions such as the ACG (Adjusted Clinical Groups) case-mix system \cite{starfield1991}, which output various mappings of the ICD codes to lower dimensional categorical spaces, some even independent of disease. A particular mapping of IDC codes to 264 Expanded Diagnosis Clusters (EDCs) was used in \cite{chang2010} to create a complex model that performed quite well in the prediction task. 

\begin{figure*}[t]
\centering
   \includegraphics[width=0.8\linewidth]{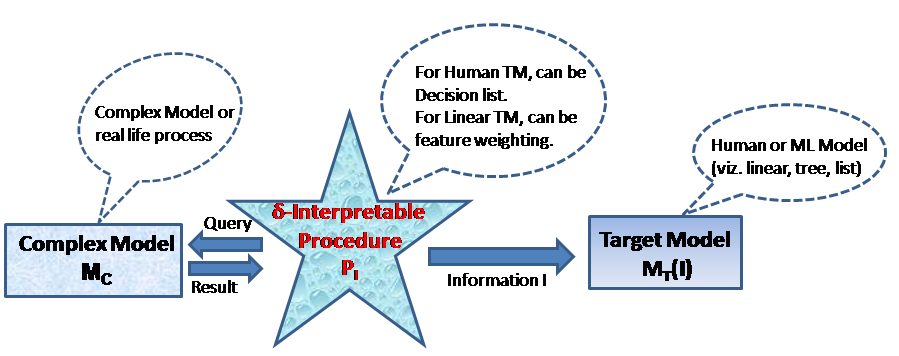}
     \caption{Above we depict what it means to be $\delta$-interpretable. Essentially, our procedure/model is $\delta$-interpretable if it improves the performance of TM by $\ge \delta$ fraction w.r.t. a target data distribution.}
\label{Intblck}
\end{figure*}


\section{Defining $\delta$-Interpretability}

Let us return to the opening question. Is interpretability simply sparsity, entropy, or something more general? An average person is said to remember no more than seven pieces of information at a time \cite{lisman1995storage}. Should that inform our notion of interpretability? Taking inspiration from the theory of computation \cite{toc} where a language is classified as regular, context free, or something else based on the strength of the machine (i.e. program) required to recognize it, we look to define interpretability along analogous lines.

Based on this discussion we would like to define interpretability relative to a target model (TM), i.e. $\delta$-interpretability. \emph{The target model in the most obvious setting would be a human, but it doesn't have to be}. It could be a linear model, a decision tree or even an entity with superhuman capabilities. The TM in our running healthcare example \cite{chang2010} is a linear model where the features come from an ACG system mapping of IDC codes to only 32 Aggregated Diagnosis Groups (ADGs). \eat{In this simple TM, the mapping is based on only five clinical features, namely; duration of condition, severity of condition, diagnostic certainty, etiology of the condition and specialty care involvement, which surprisingly do not identify organ systems or disease.}

Our model/procedure would qualify as being $\delta$-interpretable if we can somehow convey information to the TM that will lead to improving its performance (e.g., accuracy or AUC or reward) for the task at hand. Hence, the $\delta$-interpretable model has to transmit information in way that is consumable by the TM. For example, if the TM is a linear model our $\delta$-interpretable model can only tell it how to modify its feature weights or which features to consider. In our healthcare example, the authors in \cite{chang2010} need a procedure to convey information from the complex 264-dimensional model to the simple linear 32-dimensional model. Any pairwise or higher order interactions would not be of use to this model. Thus, if our "interpretable" model came up with some modifications to pairwise interaction terms, it would not be considered as an $\delta$-interpretable procedure for the linear TM. 

Ideally, the performance of the TM should improve w.r.t. to some target distribution. The target distribution could just be the underlying distribution, or it could be some reweighted form of it in case we are interested in some localities of the feature space more than others. For instance, in a standard supervised setting this could just be generalization error (GE), but in situations where we want to focus on local behavior the error would be w.r.t. the new reweighted distribution that focuses on the specific region. \emph{In other words, we allow for instance level interpretability as well as global interpretability and capturing of local behaviors that lie in between}. The healthcare example focuses on mean absolute prediction error (MAPE) expressed as a percentage of the mean of the actual expenditure (Table 3 in \cite{chang2010}). Formally, we define $\delta$-interpretability as follows:

\begin{bigboxit}
\label{didef} 
\begin{definition} $\delta$-interpretability: 
Given a target model $M_T$ belonging to a hypothesis class $\mathcal{H}$ and a target distribution $D_T$, a procedure $P_I$ is $\delta$-interpretable if the information $I$ it communicates to $M_T$ resulting in model $M_T(I)\in\mathcal{H}$ satisfies the following inequality: $e_{M_T(I)}\le \delta\cdot e_{M_T}$, where $e_{\mathcal{M}}$ is the expected error of $\mathcal{M}$ relative to some loss function on $D_T$.
\end{definition}
\end{bigboxit}

The above definition is a general notion of interpretability that does not require the interpretable procedure to have access to a complex model. It may use the complex model (CM) but it may very well act as an oracle conjuring up useful information that will improve the performance of the TM. The more intuitive but special case of Definition \ref{didef}.1 is given below which defines $\delta$-interpretability for a CM relative to a TM as being able to transfer information from the CM to the TM using a procedure $P_I$ so as to improve the TMs performance. These concepts are depicted in figure \ref{Intblck}. 

\begin{bigboxit}
\label{didefa} 
\begin{definition} CM-based $\delta$-interpretability: 
Given a target model $M_T$ belonging to a hypothesis class $\mathcal{H}$, a complex model $M_C$, and a target distribution $D_T$,
the model $M_C$ is $\delta$-interpretable relative to $M_T$, if there exists a procedure $P_I$ that derives information $I$ from $M_C$ and communicates it to $M_T$ resulting in model $M_T(I)\in\mathcal{H}$ satisfying the following inequality: $e_{M_T(I)}\le \delta\cdot e_{M_T}$, where $e_{\mathcal{M}}$ is the expected error of $\mathcal{M}$ relative to some loss function on $D_T$.
\end{definition}
\end{bigboxit}

One may consider the more intuitive definition of $\delta$-interpretability when there is a CM.

We now clarify the use of the term Information $I$ in the definition. In a normal binary classification task, training label $y \in \{+1,-1\}$ can be considered to be a one bit information about the sample $x$, i.e. "Which label is more likely given $x$?", whereas   the confidence score $p(y|x)$ holds richer information, i.e. "How likely is the label y for the sample $x$?". From an information theoretic point of view, given $x$ and only its training label $y$, there is still uncertainty about $p(y|x)$ in the interval $[1/2,1]$ prior to training. According to our definition, an interpretable method can provide useful information $I$ in the form of a sequence of bits or parameters about the training data that can potentially reduce this uncertainty of the confidence score of the TM prior to training. Moreover, the new $M_T(I)$ is better performing if it can effectively use this information.
\eat{can actually reduce the uncertainty of the $p(y(x)|x)$ in the interval $[1/2,1]$ prior to training the TM. However, the new $M_T(I)$ obtained is more useful if the training method can effectively use this potentially useful information. This is one precise and concrete sense in which, $I$ communicated by the interpretable method are indeed information bits that reduce uncertainty about the confidence score.}

\emph{Our definitions above are motivated by the fact that when people ask for an interpretation there is an implicit quality requirement in that the interpretation should be related to the task at hand. We capture this relatedness of the interpretation to the task by requiring that the interpretable procedure improve the performance of the TM. Note the TM does not have to be interpretable, rather it just is a benchmark used to measure the relevance of the provided interpretation. Without such a requirement anything that one can elucidate is then an explanation for everything else, making the concept of interpretation pointless. Consequently, the crux for any application in our setting is to come up with an interpretable procedure that can ideally improve the performance of the given TM.}

The closer $\delta$ is to 0 the more interpretable the procedure. Note the error reduction is relative to the TM model itself, not relative to the complex model. An illustration of the above definition is seen in figure \ref{Intblck}. Here we want to interpret a complex process relative to a given TM and target distribution. The interaction with the complex process could just be by observing inputs and outputs or could be through delving into the inner workings of the complex process. \emph{In addition, it is imperative that $M_T(I)\in \mathcal{H}$ i.e. the information conveyed should be within the representational power of the TM.} 

\emph{The advantage of this definition is that the TM isn't tied to any specific entity such as a human and thus neither is our definition.} We can thus test the utility of our definition w.r.t. simpler models (viz. linear, decision lists, etc.) given that a humans complexity maybe hard to characterize. We see examples of this in the coming sections.

Moreover, a direct consequence of our definition is that it naturally \emph{creates a partial ordering of interpretable procedures} relative to a TM and target distribution, which is in spirit similar to complexity classes for time or space of algorithms. For instance, if $\mathcal{R^+}$ denotes the non-negative real line $\delta_1$-interpretability $\Rightarrow$ $\delta_2$-interpretability, where $\delta_1 \le \delta_2$ $\forall \delta_1,~\delta_2\in \mathcal{R^+}$, but not the other way around. 

\section{Practical Definition of Interpretability}

We first extend our $\delta$-interpretability definition to the practical setting where we don't have the target distribution, but rather just samples. We then show how this new definition reduces to our original definition in the ideal setting where we have access to the target distribution.

\subsection{($\delta,\gamma$)-Interpretability: Performance and Robustness}

Our definition of $\delta$-interpretability just focuses on the performance of the TM. However, in most practical applications robustness is a key requirement. Imagine a doctor advising a treatment to a patient. He better have high confidence in the treatments effect before prescribing.

So what really is a robust model? Intuitively, it is a notion where one expects the same (or similar) performance from the model when applied to "nearby" inputs. In practice, this is many times done by perturbing the test set and then evaluating performance of the model \cite{carlini}. If the accuracies are comparable to the original test set then the model is deemed robust. Hence, this procedure can be viewed as creating alternate test sets on which we test the model. Thus, the procedures to create adversarial examples or perturbations can be said to induce a distribution $D_R$ from which we get these alternate test sets. \emph{The important underlying assumption here is that the newly created test sets are at least moderately probable w.r.t. the target distribution.} Of course, in case of non-uniform loss functions the test sets on whom the expected loss is low are uninteresting. This brings us to the question of when is it truly interesting to study robustness.

\emph{It seems that robustness is really only an issue when your test data on which you evaluate is incomplete i.e. it doesn't include all examples in the domain.} If you can test on all points in your domain, which could be finite, and are accurate on it then there is no need for robustness. That is why in a certain sense, low generalization error already captures robustness since the error is over the entire domain and it is impossible for your classifier to not be robust and have low GE if you could actually test on the entire domain. The problem is really only because of estimation on incomplete test sets \cite{kushtst}. \eat{Using conditional entropy definition for GE ie H(Y|X) as Karthik was suggesting doesnt solve the problem for incomplete test sets since we have seen time and again with these networks that they may give high confidences for the correct class on these test sets but are still not robust.} Given this we extend our definition of $\delta$-interpretability for practical scenarios.

\begin{bigboxit}
\begin{definition}($\delta,\gamma$)-interpretability:\label{didefp} Given a target model $M_T$ belonging to a hypothesis class $\mathcal{H}$, a sample $S_T$ from the target distribution $D_T$, a sample $S_R$ from the adversarial distribution $D_R$, a model $P_I$ is $\delta, \gamma$-interpretable relative to $(D_T\sim) S_T$ and $(D_R\sim) S_R$ if the information $I$ it communicates to $M_T$ resulting in model $M_T(I)\in\mathcal{H}$ satisfies the following inequalities:
\begin{itemize}
\item $\hat{e}^{S_T}_{M_T(I)}\le \delta\cdot \hat{e}^{S_T}_{M_T}$ (performance)
\item $\hat{e}^{S_R}_{M_T(I)}-\hat{e}^{S_T}_{M_T(I)}\le \gamma\cdot (\hat{e}^{S_R}_{M_T}-\hat{e}^{S_T}_{M_T})$ (robustness)
\end{itemize}
where $\hat{e}^{\mathcal{S}}_{\mathcal{M}}$ is the empirical error of $\mathcal{M}$ relative to some loss function.
\end{definition}
\end{bigboxit}

The first term above is analogous to the one in Definition \ref{didef}.1. The second term captures robustness and asks how representative is the test error of $M_T(I)$ w.r.t. its error on other high probability samples when compared with the performance of $M_T$ on the same test and robust sets. This can be viewed as an orthogonal metric to evaluate interpretable procedures in the practical setting. This definition could also be adapted to a more intuitive but restrictive definition analogous to Definition \ref{didefa}.2.


\subsection{Reduction to Definition \ref{didef}.1}

\emph{An (ideal) adversarial example is not just one which a model predicts incorrectly, but rather it must satisfy also the additional requirement of being a highly probable sample from the target distribution.} Without the second condition even highly unlikely outliers would be adversarial examples. But in practice this is not what people usually mean, when one talks about adversarial examples.

Given this, ideally, we should choose $D_R=D_T$ so that we test the model mainly on important examples. If we could do this and test on the entire domain our Definition \ref{didefp} would reduce to Definition \ref{didef}.1 as seen in the following proposition.

\begin{proof}[Proposition 1]
In the ideal setting, where we know $D_T$, we could set $D_R=D_T$ and compute the true errors, ($\delta,\gamma$)-interpretability would reduce to $\delta$-interpretability.
\end{proof}
\begin{proof}[Proof Sketch]
Since $D_R=D_T$, by taking expectations we get for the first condition: $E[\hat{e}^{S_T}_{M_T(I)} - \delta\hat{e}^{S_T}_{M_T}]\le 0$ and hence $e_{M_T(I)}-\delta\cdot e_{M_T}\le 0$.
For the second equation we get: $E[\hat{e}^{S_R}_{M_T(I)} - \hat{e}^{S_T}_{M_T(I)}-\gamma \hat{e}^{S_R}_{M_T}+\gamma\hat{e}^{S_T}_{M_T}]\le 0$ and hence $e_{M_T(I)}-e_{M_T(I)}-\gamma e_{M_T}+\gamma e_{M_T}\le 0$, which implies $0\le 0$.
\end{proof}

The second condition vanishes and the first condition is just the definition of $\delta$-interpretability. Our extended definition is thus consistent with Definition \ref{didef}.1 where we have access to the target distribution.

\noindent\textbf{Remark:} Model evaluation sometimes requires us to use multiple training and test sets, such as when doing cross-validation. In such cases, we have multiple target models $M^i_T$ trained on independent data sets, and multiple independent test samples $S^i_T$ (indexed by $i=\{1,\ldots,K\}$). The 
empirical error above can be defined as $(\sum_{i=1}^K{\hat{e}_{M^i_T}^{S^i_T}})/K$. Since $S^i_T$, as well as the training sets, are sampled from the same target distribution $D_T$, the reduction to Definition \ref{didef}.1 would still apply to this average error, since $E[\hat{e}_{M^h_T}^{S^i_T}] = E[\hat{e}_{M^j_T}^{S^k_T}]$ $\forall$ $h,i,j,k$.

\begin{table*}[htbp]
\begin{center}
  \begin{tabular}{|c|c|c|c|c|c|c|}
    \hline
Interpretable & TM & $\delta$ & $\gamma$ & $D_R$ & Dataset ($S_T$) & Performance\\
Procedure &&&&&& Metric\\
\hline
\hline
EDC Selection & OLS & 0.925 & 0 & Identity & Medical Claims & MAPE\\
\hline
Defensive Distillation & DNN & 1.27  & 0.8 & $L_2$ attack & MNIST & Classification error\\
\hline
MMD-critic & NPC & 0.24 & 0.98 & Skewed & MNIST & Classification error\\
\hline
LIME & SLR & 0.1 & 0 & Identity & Books & Feature Recall\\
\hline
Rule Lists (size $\le 4$ ) & Human & 0.95 & 0 & Identity & Manufacturing & Yield\\
\hline
\end{tabular}
\end{center}
  \caption{Above we see how our framework can be used to characterize interpretability of methods across applications.}
\label{inttbl}
\end{table*}

\section{Application to Existing Interpretable Methods}

We now look at how some of the current methods and how they fit into our framework.

\noindent\textbf{EDC Selection:} The running healthcare example of \cite{chang2010} considers a complex model based on 264 EDC features and a simpler linear model based on 32 ACG features, and both models also include the same demographic variables. The complex model has a MAPE of 96.52\% while the linear model has a MAPE of 103.64\%. The authors in \cite{chang2010} attempt to improve the TM's performance by including several EDC features. They develop a stepwise process for generating selected EDCs based on significance testing and broad applicability to various types of expenditures (\cite{chang2010}, Additional File 1). This stepwise process can be viewed as a $(\delta, \gamma)$-interpretable procedure that provides information in the form of 19 EDC variables which, when added to the TM, improve the performance from 103.64\% to 95.83\% and is thus (0.925, 0) -interpretable. Note the significance since, given the large population sizes and high mean annual healthcare costs per individual, even small improvements in accuracy can have high monetary impact.

\noindent\textbf{Distillation:} DNNs are not considered interpretable. However, if you consider a DNN to be a TM then you can view defensive distillation as a $(\delta, \gamma)$-interpretable procedure. We compute $\delta$ and $\gamma$ from results presented in \cite{carlini} on the MNIST dataset, where the authors adversarially perturbs the test instances by a slight amount. We see here that defensive distillation makes the DNN slightly more robust at the expense of it being a little less accurate.

\noindent\textbf{Prototype Selection:} We implemented and ran MMD-critic \cite{l2c} on randomly sampled MNIST training sets of size 1500 where the number of prototypes was set to 200. The test sets were 5420 in size which is the size of the least frequent digit. We had a representative test set and then 10 highly skewed test sets where each contained only a single digit. The representative test set was used to estimate $\delta$ and the 10 skewed test sets were used to compute $\gamma$. The TM was a nearest prototype classifier \cite{l2c} that was initialized with random 200 prototypes which it used to create the initial classifications. We see from the table that mmd-critic has a low $\delta$ and almost 1 $\gamma$ value. This implies that it is significantly more accurate than random prototype selection while maintaining robustness. In other words, it should be strongly preferred over random selection.

\noindent\textbf{LIME:} We consider the experiment in \cite{lime} where they use sparse logistic regression (SLR) to classify a review as positive or negative on the Books dataset. Their main objective here is to see if their approach is superior to other approaches in terms of selecting the true important features. There is no robustness experiment here so $D_R$ is identity which means same as $D_T$ and hence $\gamma$ is 0. Their accuracy however, in selecting the important features is significantly better than random feature selection which is signified by $\delta=0.1$. The other experiments can also be characterized in similar fashion. In cases where only explanations are provided with no explicit metric one can view as the experts confidence in the method as a metric which good explanations will enhance.

\eat{\noindent\textbf{Interpretable MDP:} We used a constrained MDP formulation \cite{imdp} to derive a product-to-product recommendation policy for the European tour operator TUI. The goal was to generate buyer conversions and to improve a simple product-to-product policy based on static pictures of the website and what products are currently looked at. The constrained MDP results in a policy that is just as simple but greatly improves the conversion rate (averaged over 10 simulations where customer behavior followed a mixed logit customer choice model with parameters fit to TUI data). The mean normalized conversion rate increased from 0.3377 to 0.6167.}

\noindent\textbf{Rule Lists:}
We built a rule list on a semi-conductor manufacturing dataset \cite{jmlr14Amit} of size 8926. In this data, a single datapoint is
a wafer, which is a group of chips, and measurements, which correspond to 1000s of input features (temperatures, pressures, gass flows, etc.), are made on this wafer throughout its production. The goal was to provide the engineer some insight into, if anything, was plaguing his process so that he can improve performance. We built a rule list \cite{twl} of size at most 4 which we showed to the engineer. The engineer figured out that there was some issue with some gas flows, which he then fixed. This resulted in 1\% more of his wafers ending up within specification. Or his yield increased from 80\% to 81\%. This is significant since, even small increase in yield corresponds to billions of dollars in savings.

\section{Framework Generalizability}
\label{assump}

It seems that our definition of $\delta$-interpretability requires a predefined goal/task. While (semi-)supervised settings have a well-defined target, we discuss other applicable settings.


In unsupervised settings although we do not have an explicit target there are quantitative measures \cite{charubook} such as Silhouette or mutual information that are used to evaluate clustering quality. Such measures which people use to evaluate quality would serve as the target loss that the $\delta$-interpretable procedure would aim to improve upon. The target distribution in this case would just be the instances in the dataset. If a generative process is assumed for the data, then that would dictate the target distribution.

In other settings such as reinforcement learning (RL) \cite{reinf}, the $\delta$-interpretable procedure would try to increase the expected discounted reward of the agent by directing it into more lucrative portions of the state space.\eat{ In inverse RL on the other hand, it would assist in learning a more accurate reward function based on the observed behaviors of an intelligent entity. The methodology could also be used to test interpretable models on how well they convey the causal structure \cite{pearl} to the TM by evaluating the TMs performance on counterfactuals before and after the information has been conveyed.}

There maybe situations where a human TM may not want to take any explicit actions based on the insight conveyed by a $\delta$-interpretable procedure. However, an implicit evaluation metric, such as human satisfaction, probably exists, which a good interpretable procedure will help to enhance. For example, in the setting where you want explanations for classifications of individual instances \cite{lime}, the metric that you are enhancing is the human confidence in the complex model.

\section{Related Work}

There has been a great deal of interest in interpretable modeling recently and for good reason. Almost any practical application with a human decision maker interpretability is imperative for the human to have confidence in the model. It has also become increasingly important in deep neural networks given their susceptibility to small perturbations that are humanly unrecognizable \cite{carlini,gan}. 

There have been multiple frameworks and algorithms proposed to perform interpretable modeling. These range from building rule/decision lists \cite{decl,twl} to finding prototypes \cite{l2c} to taking inspiration from psychometrics \cite{irt} and learning models that can be consumed by humans. There are also works \cite{lime} which focus on answering instance specific user queries by locally approximating a superior performing complex model with a simpler easy to understand one which could be used to gain confidence in the complex model.

The most relevant work to our current endeavor is possibly \cite{rsi}. They provide an in depth discussion for why interpretability is needed, and an overall taxonomy for what should be considered when talking about interpretability. Their final TM is always a human even for the functionally grounded explanation as the TMs are proxies for human complexity. As we have seen, our definition of a TM is more general, as besides human, it could be any ML model or even something else that has superhuman cognitive skills. This generalization allows us to test our definition without the need to pin down human complexity. Moreover, we provide a formal strict definition for $\delta$-interpretability that accounts for key concepts such as performance and robustness and articulates how robustness is only an issue when talking about incomplete test sets.\eat{ In addition we also propose a principled meta-interpretable strategy that works well in practice. Our meta strategy has relations to distillation and learning with privileged information \cite{priv16} with the key difference being in the mechanics of how we use information which is by weighting the instances rather than modeling it as a target. This has the advantage of not having to change from a classification to regression setting. Moreover, weighting instances has an intuitive justification where if you view the complex model as a teacher and the TM as a student, the teacher is telling the student which aspects (e.g. instances) he should focus on and which he could ignore.}

\section{Discussion}

We defined $\delta$ for a single distribution but it could be defined over multiple distributions where $\delta = \max(\delta_1,...,\delta_k)$ for $k$ distributions and analogously $\gamma$ also could be defined over multiple adversarial distributions. We did not include these complexities in the definitions so as not to lose the main point, but extensions such as these are straightforward.

Another extension could be to have a sensitivity parameter $\alpha$ to define equivalence classes, where if two models are $\delta_1$- and $\delta_2$-interpretable, then they are in the same equivalence class if $|\delta_1-\delta_2|\le \alpha$. This can help group together models that can be considered to be equivalent for the application at hand. The $\alpha$ in essence quantifies operational significance. One can have even multiple $\alpha$ as a function of the $\delta$ values.

One can also extend the notion of interpretability where $\delta$ or and $\gamma$ are 1 but you can learn the same model with fewer samples given information from the interpretable procedure.

Human subjects store approximately 7 pieces of information \cite{lisman1995storage}. As such, we can explore highly interpretable models, which can be readily learned by humans, by considering models for TM that make simple use of no more than 7 pieces of information. Feldman~\cite{feldman2000minimization} finds that the subjective difficulty of a concept is directly proportional to its Boolean complexity, the length of the shortest logically equivalent propositional formula. We could explore interpretable models of this type. Another model bounds the rademacher complexity of humans \cite{zhu2009human} as a function of complexity of the domain and sample size. Although the bounds are loose, they follow the empirical trend seen in their experiments on words and images.

Finally, all humans may not be equal relative to a task. Having expertise in a domain may increase the level of detail consumable by that human. So the above models which try to approximate human capability may be extended to account for the additional complexity consumable by the human depending on their experience.

\section*{Acknowledgements}
We would like to thank Margareta Ackerman, Murray Campbell, Alexandra Olteanu, Marek Petrik, Irina Rish, Kush Varshney and Bowen Zhou for insightful suggestions and comments.


\bibliography{FFInterpret}
\bibliographystyle{icml2017}

\end{document}